\pdfoutput=1

\documentclass[english]{article} % For LaTeX2e
\usepackage[nonatbib,arxiv]{nips_2016}
\usepackage[numbers]{natbib}
\usepackage{times}
\usepackage{url}
\usepackage{graphicx}
\usepackage{babel}
\usepackage{amsmath}
\usepackage{amsthm}
\usepackage{multirow}
\usepackage{subcaption}
\usepackage[ruled,linesnumbered]{algorithm2e}
\usepackage{wrapfig}
\usepackage{enumitem}
\usepackage{xcolor}
\newtheorem{thm}{\protect\theoremname}
\theoremstyle{definition}
\newtheorem{defn}[thm]{\protect\definitionname}
\providecommand{\definitionname}{Definition}
\theoremstyle{lemma}

\newtheorem*{prop*}{\protect\propositionname}
\newtheorem{prop}{\protect\propositionname}
\providecommand{\lemmaname}{Lemma}
\providecommand{\theoremname}{Theorem}
\providecommand{\propositionname}{Proposition}

\DeclareMathOperator*{\argmax}{argmax}

\SetKwInput{KwParameter}{Parameter}
\SetKwInput{KwInput}{Input}
\SetKwRepeat{Do}{do}{while}

\title{Logarithmic Time One-Against-Some}

\author{
Hal Daum\'e III \\
University of Maryland \\
\texttt{me@hal3.name} \And
Nikos Karampatziakis \\
Microsoft \\
\texttt{nikosk@microsoft.com} \AND
John Langford \\
Microsoft \\
\texttt{jcl@microsoft.com} \And
Paul Mineiro \\
Microsoft \\
\texttt{pmineiro@microsoft.com} 
}

\begin{document}

\maketitle
\begin{abstract}
We create a new online reduction of multiclass classification to
binary classification for which training and prediction time scale
logarithmically with the number of classes. We show that several
simple techniques give rise to an algorithm that can compete with
one-against-all in both space and predictive power while offering
exponential improvements in speed when the number of classes is large.
\end{abstract}

\section{Introduction}

Can we effectively predict one of $K$ classes in polylogarithmic 
time in $K$?  This question gives rise to the area of extreme multiclass classification~\cite{bengio2010label,FilterTree,bhatia2015locally,Lomtree,morin2005hierarchical,prabhu2014fastxml,weston2013label}, in which $K$ is very large.
If efficiency is not a concern, the most common and generally effective
representation for multiclass prediction is a one-against-all (OAA) structure.
Here, inference consists of computing a score for each class and returning 
the class with the maximum score. 
An attractive strategy for picking one of $K$ items efficiently is to use a tree; unfortunately, this often comes at the cost of increased error.

A general replacement for the one-against-all approach must satisfy
a difficult set of desiderata.
\begin{itemize}[noitemsep,nolistsep]
\item High accuracy: The approach should provide accuracy competitive
  with OAA, a remarkably strong baseline\cite{rifkin2004defense} which
  is the standard ``output layer'' of many learning
  systems such as winners of the ImageNet contest~\cite{resnet,VGG}.
\item High speed at training time \emph{and} test time: A multiclass
  classifier must spend at least $\Omega(\log K)$ time~\cite{Lomtree})
  so this is a natural benchmark to optimize against.
\item Online operation:  Many learning algorithms use either online
  updates or mini-batch updates.  Approaches satisfying this
  constraint can be easily composed into an end-to-end learning system
  for solving complex problems like image recognition. For algorithms
  which operate in batch fashion, online components can be easily
  used.
\item Linear space: In order to have a drop-in replacement for OAA, an
  approach must not take much more space than OAA.  Memory is at a
  premium when $K$ is very large, especially for models trained on
  GPUs, or deployed to small devices.
\end{itemize}

%Now we talk about the form of our solution.
We use an OAA-like structure to make a final prediction, but instead
of scoring \emph{every} class, we only score a small subset of $O(\log
K)$ classes.  We call this ``one-against-some'' (OAS).  How can you
efficiently determine what classes should be scored?  We use a
\emph{dynamically} built tree to efficiently whittle down the set of
candidate classes.  The goal of the tree is to maximize the
\emph{recall} of the candidate set so we call this approach
``The Recall Tree.''

\begin{figure}
  \hspace{1em}
\begin{subfigure}{0.4\textwidth}
\begin{center}
  \begin{algorithm}[H]
    \TitleOfAlgo{Recall\_Tree\_Test}
\KwInput{Example $x$; Root Node $\mathrm{n}$}
\KwResult{Predicted class $\hat y$}
\Do{n.leaf is false}{
$\mathrm{r} \leftarrow f_\mathrm{n}(x) > 0$ ? n.left : n.right \;
\If{$\widehat{\mathbf{recall}} (\mathrm{n}) > \widehat{\mathbf{recall}} (\mathrm{r})$}{
\textbf{break}
}
$\mathrm{n} \leftarrow \mathrm{r}$ \;
$x \leftarrow x \wedge \{ (n:1) \}$
}
$\hat y \leftarrow \argmax\limits_{y \in \mathrm{n.candidates}} \mbox{Predict}_y(x)$
\end{algorithm}
\end{center}
\end{subfigure}
\begin{subfigure}{0.49 \textwidth}
\begin{center}
\hspace{-8ex}
\includegraphics[width=0.9\textwidth,trim={248mm 115mm 155mm 17mm},clip]{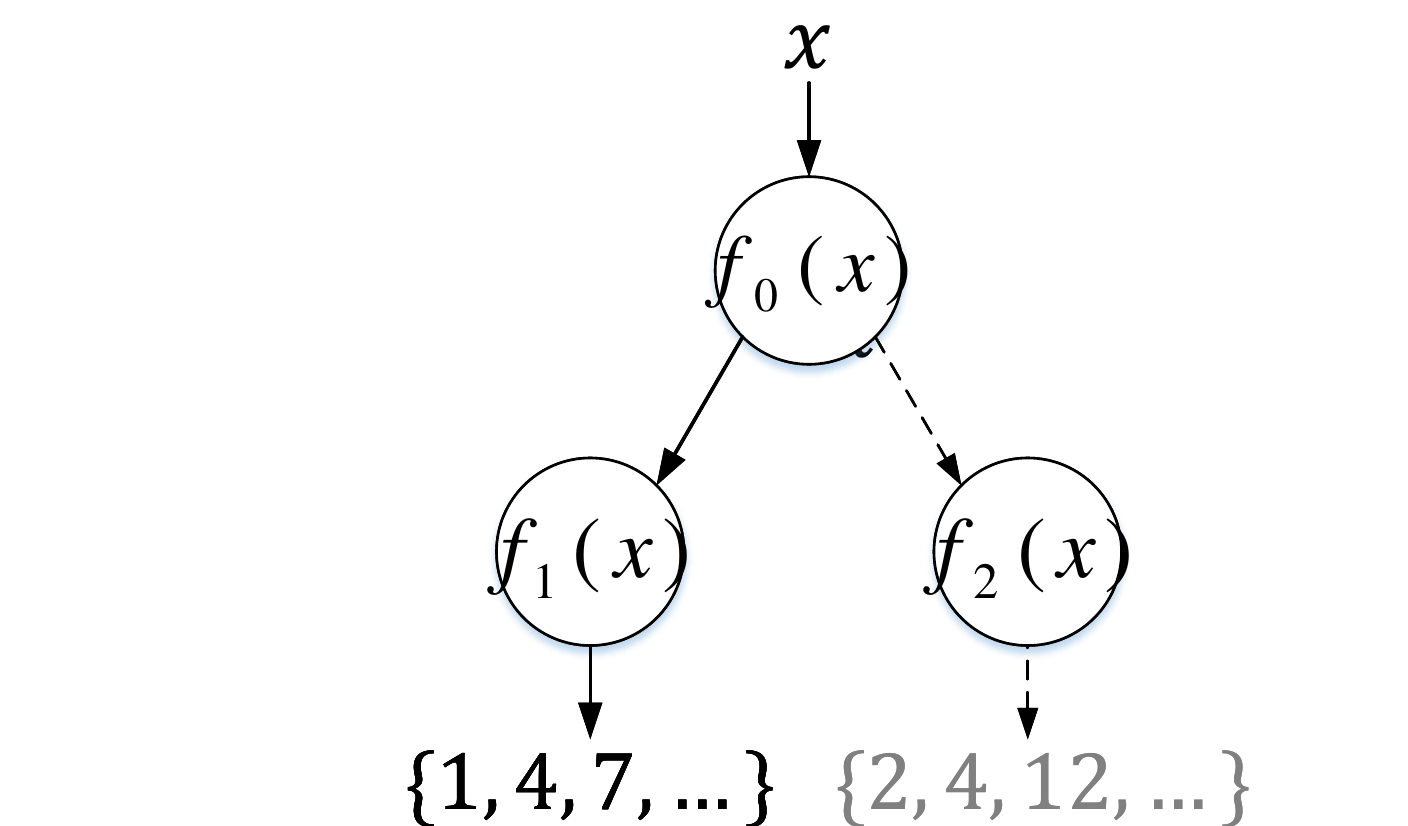}
\end{center}
\end{subfigure}
\caption{Left: Pseudocode for prediction where $f_\mathrm{n}(x)$ evaluates the node's route, $\mbox{Predict}_y(x)$ evaluates a per-class regressor, $\widehat{\mathbf{recall}} (\circ)$ is an empirical bound on the recall of a node $(\circ)$ (see section~\ref{sec:recall}), and $x \wedge \{ (n:1) \}$ indicates the addition of a sparse feature with index $n$ and value $1$.  Right: An example is routed through the tree to a leaf node associated with a set of eligible classes. }
\label{fig:inference}
\end{figure}

Figure~\ref{fig:inference} depicts the inference procedure for the
Recall Tree: an example is routed through a tree until termination,
and then the set of eligible classes compete to predict the label.  We
use this inference procedure at training time, to facilitate
end-to-end joint optimization of the predictors at each internal node
in the tree (the ``routers''), the tree structure, and the final OAS
predictors.

%Now we talk about its properties.
The Recall Tree achieves good accuracy, improving on previous
online approaches~\cite{Lomtree} and sometimes surpassing the OAA
baseline.  The algorithm requires only $\mathrm{poly}(\log
K)$ time during training and testing.  In practice, the computational
benefits are substantial when $K\geq 1000$.\footnote{Our
  implementation of baseline approaches, including OAA, involve
  vectorized computations that increase throughput by a factor of $10$
  to $20$, making them much more difficult to outpace than na\"ive
  implementations.} The Recall Tree constructs a tree and learns
parameters in a fully online manner as a reduction allowing
composition with systems trained via online updates.  All of this
requires only a factor of 2 more space than OAA approaches.
 
Our contributions are the following:
\begin{itemize}[noitemsep,nolistsep]
\item We propose a new online tree construction algorithm which
  jointly optimizes the construction of the tree, the routers and the
  underlying OAS predictors (see section~\ref{sec:recall}).
\item We analyze elements of the algorithm, including a new
  boosting bound (see section~\ref{sec:boosting}) on multiclass
  classification performance and a representational trick which allows
  the algorithm to perform well if \emph{either} a tree representation
  does well or a OAA representation does well as discussed in
  section~\ref{sec:path}. 
\item We experiment with the new algorithm, both to analyze its
  performance relative to baselines and understand the impact of
  design decisions via ablation experiments.
\end{itemize}
The net effect is a theoretically motivated algorithm which
empirically performs well providing a plausible replacement for the
standard one-against-all approach in the large $K$ setting.

\subsection{Prior Work}

The LOMTree\cite{CCB16,Lomtree} is the closest prior work.  It misses
on space requirements: up to a factor of 64 more space than OAA was
used experimentally.  Despite working with radically less space we
show the Recall Tree typically provides better predictive performance.
The key differences here are algorithmic: A tighter reduction at
internal nodes and the one-against-some approach yields generally
better performance despite much tighter resource constraints.

Boosting trees~\cite{KMBoosting} for multiclass learning~\cite{TM03}
on a generalized notion of entropy are known to results in low 0/1
loss.  Relative to these works we show \emph{how} to efficiently
achieve weak learning by reduction to binary classification making
this approach empirically practical.  We also address a structural
issue in the multiclass analysis (see section~\ref{sec:boosting}).

Other approaches such as hierarchical softmax (HSM) and the the Filter
Tree~\cite{FilterTree} use a fixed tree
structure~\cite{morin2005hierarchical}.  In domains in which there is
no prespecified tree hierarchy, using a random tree structure can lead
to considerable underperformance as shown
previously~\cite{bengio2010label,Lomtree}.

Most other approaches in extreme classification either do not work
online \cite{mnih2009scalable,prabhu2014fastxml} or only focus on
speeding up either prediction time or training time but not both.
Most of the works that enjoy sublinear inference time (but
(super)linear training time) are based on tree decomposition
approaches. In \cite{mnih2009scalable} the authors try to add tree
structure learning to HSM via iteratively clustering the
classes. While the end result is a classifier whose inference time
scales logarithmically with the number of classes, the clustering
steps are batch and scale poorly with the number of classes. Similar
remarks apply to \cite{bengio2010label} where the authors propose to
learn a tree by solving an eigenvalue problem after (OAA)
training. The work of~\cite{weston2013label} is similar in spirit to
ours, as the authors propose to learn a label filter to reduce the
number of candidate classes in an OAA approach. However they learn the
tree after training the underlying OAA predictors while here
we learn and, more crucially, use the tree during training of the OAS
predictors. Among the approaches that speed up training time we
distinguish exact ones~\cite{de2015exploration, NIPS2015_5865} that
have only been proposed for particular loss functions and approximate
ones such as negative sampling as used
e.g.\ in~\cite{weston2011wsabie}. Though these techniques do not
address inference time, separate procedures for speeding up inference
(given a trained model) have been proposed
\cite{shrivastava2014asymmetric}. However, such two step procedures
can lead to substantially suboptimal results.

\section{The Recall Tree Algorithm}

Here we present a concrete description of the Recall Tree and defer
all theoretical results that motivate our decisions to the next
section.  The Recall Tree maintains one predictor for each class and a
tree whose purpose is to eliminate predictors from consideration.  We
refer to the per-class predictors as one-against-some (OAS)
predictors.  The tree creates a high recall set of candidate classes
and then leverages the OAS predictors to achieve precision.
Crucially, the leaves of the tree do \emph{not} partition the set of
classes: classes can (and do) have support at multiple leaves.

\begin{figure}[t]
\hspace{1em}
\begin{subfigure}{0.46 \textwidth}
\begin{algorithm}[H]
\TitleOfAlgo{Recall\_Tree\_Train}
\KwInput{Example $(x, y)$; Root node $n$}
\KwResult{Updated tree with root at $n$}
\Do{n.leaf is false}{
$\mathbf{update\_router} (x, y, \mathrm{n})$ \;
$\mathrm{r} \leftarrow f_\mathrm{n} (x) > 0$ ? n.left : n.right \;
$\mathbf{update\_candidates} (x, y, \mathrm{r})$ \;
\If{$\widehat{\mathbf{recall}} (\mathrm{n}) > \widehat{\mathbf{recall}} (\mathrm{r})$}{
\textbf{break}
}
$\mathrm{n} \leftarrow \mathrm{r}$\;
$x \leftarrow x \cup \{ (n:1) \}$
}
$\mathbf{update\_predictors} (x, y, \mathrm{n.candidates})$ \;
\end{algorithm}
\caption{An input labeled example descends the tree updating routers
  at the nodes until reaching a leaf or a recall estimate declines,
  leading to an early break.  Here, $\mathbf{update\_candidates}$
  updates the set of candidate labels at each node and
  $\mathbf{update\_predictors}$ updates the one-against-some
  predictors upon breaking; and $\widehat{\mathbf{recall}} (\circ)$ is a an empirical bound on the recall of a node $(\circ)$ (see section~\ref{sec:recall}).}
\label{subfig:learning}
\end{subfigure}
\hspace{0.03\textwidth}
\begin{subfigure}{0.46 \textwidth}
\begin{algorithm}[H]
\TitleOfAlgo{update\_router}
  \KwInput{Example $(x, y)$; Node $n$}
\KwResult{Update node $n$}
$\hat H_{\mathrm{left}}, \hat H_{\mathrm{left}}' \doteq \mathbf{entropy} (\mathrm{n.left},y)$ \;
$\hat H_{\mathrm{right}}, \hat H_{\mathrm{right}}' \doteq \mathbf{entropy} (\mathrm{n.right},y)$ \;
$\hat H_{\mathrm{|left}} \doteq \frac{\mathrm{n.left.total}}{\mathrm{n.total}} \hat H_{\mathrm{left}}' +\frac{\mathrm{n.right.total}}{\mathrm{n.total}} \hat H_{\mathrm{right}}$ \;
$\hat H_{\mathrm{|right}} \doteq \frac{\mathrm{n.left.total}}{\mathrm{n.total}} \hat H_{\mathrm{left}} +\frac{\mathrm{n.right.total}}{\mathrm{n.total}} \hat H_{\mathrm{right}}'$ \;
$\widehat{\Delta H}_{\mathrm{post}} \leftarrow \hat H_{\mathrm{|left}} -\hat H_{\mathrm{|right}}$  \;
$\mathrm{Learn}_n (x, | \widehat{\Delta H}_{\mathrm{post}}|, \mathrm{sign} (\widehat{\Delta H}_{\mathrm{post}}))$
\end{algorithm}
\caption{Here, $\mathbf{entropy}$ computes
  the empirical entropy of labels incident on a node without and with
  (respectively) an extra label $y$.  $\hat H_{\mathrm{|left}}$ is an estimate of the
  average entropy if the example is routed left.  $\mathrm{Learn}_n
  (x, w, y)$ is an importance-weighted update to the binary classifier
  $f(x)$ for node $n$ with features $x$, label $y$, and weight $w$.}
\label{subfig:updaterouter}
\end{subfigure} \\
\caption{Learning procedure.}
\label{fig:learning}
\end{figure}

Figure~\ref{fig:learning} outlines the learning procedures, which we now describe in more detail.
Each node in the tree maintains a set of statistics. % that are necessary for estimation.
First, each node $n$ maintains a \emph{router}, denoted $f_\mathrm{n}$, that maps an example to either a left or right child. This router is implemented as a binary classifier. Second, each node maintains a histogram of the labels of all training examples that have been routed to, or through, that node. This histogram is used in two ways: (1) the most frequent classes form the competitor set for the OAS predictors; (2) the histogram is used to
decide whether the statistics at each node can be trusted. This is a crucial issue with trees because a child node sees fewer data than its parent. Therefore we do not simply
rely on the \emph{empirical recall} (i.e.\ the observed fraction of labels that fall into the most frequent $F$ labels at this node) of a node since such estimate can have considerable variance at deep nodes. Instead, we use a lower bound of the true recall which we compute via an empirical Bernstein inequality (see Section~\ref{sec:recall}). 

\textbf{Learning the predictors for each class}  In Figure~\ref{subfig:learning} $\mathbf{update\_predictors}$ updates the candidate set predictors using the standard OAA strategy restricted to the set of eligible classes.  If the true label is not in the $F$ most frequent classes at this node then no update occurs.

\textbf{Learning the set of candidates in each node}  In Figure~\ref{subfig:learning} $\mathbf{update\_candidates}$  updates the count of the true label at this node.
At each node, the most frequent $F$ labels are the candidate set.

\textbf{Learning the routers at each node}  In Figure~\ref{subfig:updaterouter} $\mathbf{update\_router}$ updates the router at a node by optimizing the decrease in the entropy of the label distribution (the label entropy) due to routing. This is in accordance with our theory (Section~\ref{sec:boosting}).  The label entropy for a node is estimated using the empirical counts of each class label entering the node. These counts are reliable as $\mathbf{update\_router}$ is only called for the root or nodes whose true recall bound is better than their children. The expected label entropy after routing is estimated by averaging the estimated label entropy of each child node, weighted by the fraction of examples routing left or right.  Finally, we compute the advantage of routing left vs.\ right by taking the difference of the expected label entropies for routing left vs.\ right. The sign of this difference determines the binary label for updating the router.  

\textbf{Tree depth control} \label{para:structure} We calculate a
lower bound $\widehat{\mathbf{recall}} (\mathrm{n})$ on the true
recall of node $n$ (Section~\ref{sec:recall}), halting descent as in
Figure~\ref{subfig:learning}. As we descend the tree, the bound first
increases (empirical recall increases) then declines (variance
increases).  We also limit the maximum depth $d$ of the tree.  This
parameter is typically not operative but adds an additional safety
check and sees some use on datasets where multipasses are employed.

\section{Theoretical Motivation} 

Online construction of an optimal logarithmic time predictor for
multiclass classification given an arbitrary fixed representation at
each node appears deeply intractable.  A primary difficulty is that
decisions have to be \emph{hard} since we cannot afford to maintain a
distribution over all class labels.  Choosing a classifier so as to
minimize error rate has been considered for cryptographic
primitives~\cite{HardBlum} so it is plausibly hard on averager rather
than merely hard in the worst case.  Furthermore, the \emph{joint}
optimization of all predictors does not nicely decompose into
independent problems. Solving the above problems requires an
implausible break-through in complexity theory which we do not achieve
here.  Instead, we use learning theory to assist the design by
analyzing various simplifications of the problem.

\subsection{One-Against-Some Prediction and Recall}
\label{sec:recall}

For binary classification, branching programs~\cite{MMBoosting} result
in exponentially more succinct representations than decision
trees~\cite{KMBoosting} by joining nodes to create directed acyclic
graphs.  The key observation is that nodes in the same level with a
similar distribution over class labels can be joined into one node,
implying that the number of nodes at one level is only
$\theta(1/\gamma)$ where $\gamma$ is the weak learning parameter
rather than exponential in the depth.  This approach generally fails
in the multiclass setting because covering the simplex of multiclass
label distributions requires $(K-1)^{\theta(1/\gamma)}$ nodes.

One easy special case exists.  When the distribution over class labels
is skewed so one label is the majority class, learning a minimum
entropy classifier is equivalent to predicting whether the class is
the majority or not.  There are only $K$ possible OAS
predictors of this sort so maintaining one for each class label is
computationally tractable.  

Using OAS classifiers creates a limited branching program
structure over predictions.  Aside from the space savings generated,
this also implies that nodes deep in the tree use many more labeled
examples than are otherwise available.  In finite sample regimes,
which are not covered by these boosting analyses, having more labeled
samples implies a higher quality predictor as per standard sample
complexity analysis.

A fundamental issue with a tree-structured prediction is that the
number of labeled examples incident on the root is much larger than
the number of labeled examples incident on a leaf.  This potentially
leads to: (1) underfitting toward the leaves; and (2) insufficient
representation complexity toward the root.  Optimizing recall, rather
than accuracy, ameliorates this drawback.  Instead of halting at a
leaf, we can halt at an internal node $n$ for which the top $F$ most
frequent labels contain the true answer with a sufficiently high
probability.  When $F = O(\log K)$ this does not compromise the goal
of achieving logarithmic time classification.

Nevertheless, as data gets divided down the branches of the tree,
empirical estimates for the ``top $F$ most frequent labels'' suffer from a
substantial missing mass problem~\cite{GoodTuring}.
Thus, instead of computing \emph{empirical recall} to determine when to
halt descent, we use an empirical Bernstein (lower) bound \cite{maurer2009empirical}, which is summarized by
the following proposition.

\begin{prop}\label{prop:bernstein}
For all learning problems $D$ and all nodes $n$ in a fixed tree there
exists a constant $\lambda > 0$ such that with probability $1-\delta$:
\begin{align}\label{eqn:bernstein}
r_n \geq \hat{r}_n - \sqrt{\frac{\lambda \hat{r}_n (1 - \hat{r}_n)}{m_n}} - \frac{\lambda}{m_n}
\end{align}
where $\hat{r}_n$ is the empirical frequency amongst $m_n$ events that
the true label is in the top $F$ labels and $r_n$ is the expected
value in the population limit.
\end{prop}

Reducing the depth of the tree by using a bound on $r_n$ and
joining labeled examples from many leaves in a one-against-some
approach both relieves data sparsity problems and allows greater
error tolerance by the root node.

\subsection{Path Features}\label{sec:path}

\begin{figure}
\begin{subfigure}{0.45\textwidth}
\includegraphics[width=\textwidth]{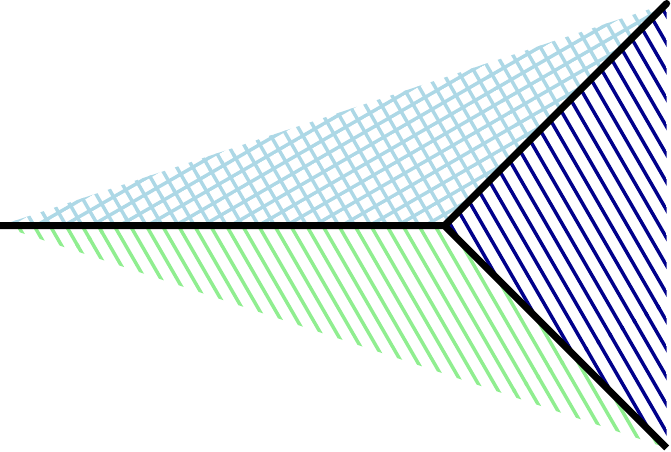}
\end{subfigure}
\hspace{0.05\textwidth}
\begin{subfigure}{0.45\textwidth}
\includegraphics[width=\textwidth]{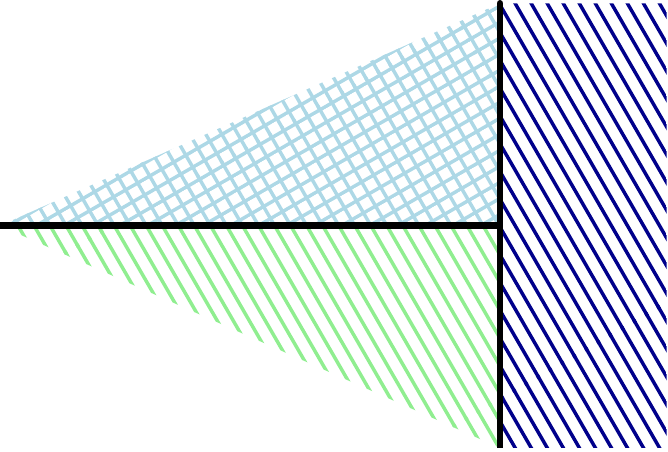}
\end{subfigure}
\caption{Two different distributions over class labels in the plane
  with each color/pattern representing support for a single class
  label.  The left distribution is easily solved with an
  OAA classifier while the right distribution is easily
  solved with a decision tree.}
\label{fig:representation}
\end{figure}

The relative representational power of different solutions is an
important consideration.  Are OAA types of representations
inherently more or less powerful than a tree based representation?
Figure~\ref{fig:representation} shows two learning problems
illustrating two extremes under the assumption of a linear
representation.

{\bf Linear OAA:} If all the class parameter vectors happen to have
the same magnitude then OAA classification is equivalent to finding
the nearest neighbor amongst a set of vectors (one per class) which
partition the space into a Voronoi diagram as
in~\ref{fig:representation} on the left.  The general case, with
unequal vectors corresponds to a weighted Voronoi diagram where the
magnitude of two vectors sharing a border determines the edge of the
partition.  No weighted Voronoi diagram can account for the partition
on the right.

{\bf Trees:} If the partition of a space can be represented by a
sequence of conditional splits, then a tree can represent the solution
accurately as in~\ref{fig:representation} on the right.  On the other
hand, extra work is generally required to represent a Voronoi diagram
as on the left.  In general, the number of edges in a multidimensional
Voronoi diagram may grow at least quadratically in the number of
points implying that the number of nodes required for a tree to
faithfully represent a Voronoi diagram is at least $\Theta(n^2)$.

Based on this, neither tree-based nor OAA style prediction
is inherently more powerful, with the best solution being problem
dependent.  

Since we are interested in starting with a tree-based approach and
ending with a OAS classifier there is a simple
representational trick which provides the best of both worlds.  We can
add features which record the path through the tree.  To be precise,
let $T$ be a tree and $\mbox{path}_T(x)$ be a vector with one
dimension per node in $T$ which is set to $1$ if $x$ traverses the
node and $0$ otherwise.  The following proposition holds.

\begin{prop*}
For any learning problem $D$ for which a tree $T$ achieves error rate
$\epsilon$, $\mbox{OAA}(x,\mbox{path}_T(x))$ with a linear
representation can achieve error rate $\epsilon$.
\end{prop*}

Linear representations are special, because they are tractably
analyzed and because they are the fundamental building blocks around
which many more complex representations are built.  Hence, this
representational change eases prediction in many common settings.

\begin{proof}
A linear OAA classifier is defined by a matrix $w_{iy}$
where $i$ ranges over the input and $y$ ranges over the labels.  Let
$w_{iy} = 0$ by default and $1$ when $i$ corresponds to a leaf for
which the tree predicts $y$.  Under this representation, the
prediction of $\mbox{OAA}(x,\mbox{path}_{T}(x))$ is identical to
$T(x)$, and hence achieves the same error rate.
\end{proof}

\subsection{Optimization Objective}\label{sec:boosting}

The Shannon Entropy of class labels is optimized in the router of
figure~\ref{subfig:updaterouter}.  Why?

Since the Recall Tree jointly optimizes over many base learning
algorithms, the systemic properties of the joint optimization are
important to consider.  A theory of decision tree learning as
boosting~\cite{KMBoosting} provides a way to understand these joint
properties in a population limit (or equivalently on a training set
iterated until convergence).  In essence, the analysis shows that each
level of the decision tree boosts the accuracy of the resulting tree
with this conclusion holding for several common objectives.

In boosting for multiclass classification~\cite{CCB16, Lomtree, TM03},
it is important to achieve a weak dependence on the number of class
labels.  Shannon Entropy is particularly well-suited to this goal,
because it has only a logarithmic dependence on the number of class
labels.  Let $\pi_{i|n}$ be the probability that the correct label is
$i$, conditioned on the corresponding example reaching node $n$. Then
$H_{n}=\sum_{i=1}^{K}\pi_{i|n}\log_2 \frac{1}{\pi_{i|n}}$ is the Shannon
entropy of class labels reaching node $n$.

For this section, we consider a simplified algorithm which neglects
concerns of finite sample analysis, how optimization is done, and the
leaf predictors.  What's left is the value of optimizing the router
objective.  We consider an algorithm which recursively splits the leaf
with the largest fraction $f$ of all examples starting at the root and
reaching the leaf.  The leaf is split into two new leaves to the left
$l$ and right $r$.  If $f_l$ and $f_r$ are the fraction of examples
going left and right, the split criterion minimizes the expectation
over the leaves of the average class entropy, $f_l H_l+f_r H_r$. This
might be achieved by $\mathbf{update\_router}$ in
Figure~\ref{subfig:learning} or by any other means.  With this
criterion we are in a position to directly optimize information
boosting.  

\begin{defn}
($\gamma$-Weak Learning Assumption) For all distributions $n(x,y)$ a
  learning algorithm using examples $(x,y)^*$ IID from $n$ finds a
  binary classifier $c:X\rightarrow\{l,r\}$ satisfying
\[
f_l H_l + f_r H_r\leq H_n-\gamma\,\,\,\,.
\]
\end{defn}
This approach is similar to previous~\cite{TM03} except that we boost
in an \emph{additive} rather than a \emph{multiplicative} sense.
This is good because it suppresses an implicit dependence
on $K$ (since for any nontrivial $\gamma$ there exists a $K$ such that
with a uniform distribution $U$, $H_U(1-\gamma) > 1$), yeilding a strictly stronger result.

As long as Weak Learning occurs, we can prove the following theorem.
\begin{thm}\label{thm:boost} If $\gamma$ Weak Learning holds for every node in the tree and nodes with the largest fraction of examples are split first, then after $t>2$ splits the multiclass error rate $\epsilon$ of the tree is bounded by:
  \[ \epsilon \leq H_1 - \gamma (1 + \ln t ) \]
where $H_1$ is the entropy of the marginal distribution of class labels.
\end{thm} 
The proof in appendix~\ref{sec:boostproof} reuses techniques
from~\cite{Lomtree, KMBoosting} but has a tighter result.

The most important observation from the theorem is that as $t$ (the
number of splits) increases, the error rate is increasingly bounded.
This rate depends on $\ln t$ agreeing with the intuition that boosting
happens level by level in the tree.  The dependence on the initial
entropy $H_1$ shows that skewed marginal class distributions are
inherently easier to learn than uniform marginal class distributions,
as might be expected.  These results are similar to previous
results~\cite{CCB16, Lomtree, KMBoosting, TM03} with advantages.  We
handle multiclass rather than binary classification~\cite{KMBoosting},
we bound error rates instead of entropy~\cite{CCB16, Lomtree}, and we
use additive rather than multiplicative weak learning~\cite{TM03}.

\section{Empirical Results}

We study several questions empirically.  
\begin{enumerate}[noitemsep,nolistsep]
\item What is the benefit of using one-against-some on a recall set?
\item What is the benefit of path features?
\item Is the online nature of the Recall Tree useful on nonstationary problems?
\item How does the Recall Tree compare to one-against-all statistically and computationally?
\item How does the Recall Tree compare to LOMTree statistically and computationally?
\end{enumerate}
Throughout this section we conduct experiments using learning with a
linear representation.

\subsection{Datasets}

\begin{table}[h]
\caption{Datasets used for experimentation.}
\label{tab:datasets}
\begin{center}
\begin{tabular}{|c|c|r|r|}
\hline Dataset & Task & Classes & Examples \\
\hline ALOI\cite{geusebroek2005amsterdam} & Visual Object Recognition & $1k$ & $10^5$ \\
Imagenet\cite{Oquab14} & Visual Object Recognition & $\approx 20k$ & $\approx 10^7$ \\
LTCB\cite{mahoney2009large} & Language Modeling & $\approx 80k$ & $\approx 10^8$ \\
ODP\cite{bennett2009refined} & Document Classification & $\approx 100k$ & $\approx 10^6$ \\
\hline
\end{tabular}
\end{center}
\end{table}

Table~\ref{tab:datasets} overviews the data sets used for experimentation.
These include the largest datasets where published results are
available for LOMTree (Aloi, Imagenet, ODP), plus an additional language
modeling data set (LTCB).  Implementations of the learning algorithms,
and scripts to reproduce the data sets and experimental results, are
available at (url redacted).  Additional details about the datasets can be found
in Appendix~\ref{app:datasets}.

\subsection{Comparison with other Algorithms}

\begin{figure}
\begin{subfigure}[t]{0.49\textwidth}
\includegraphics[width=\textwidth]{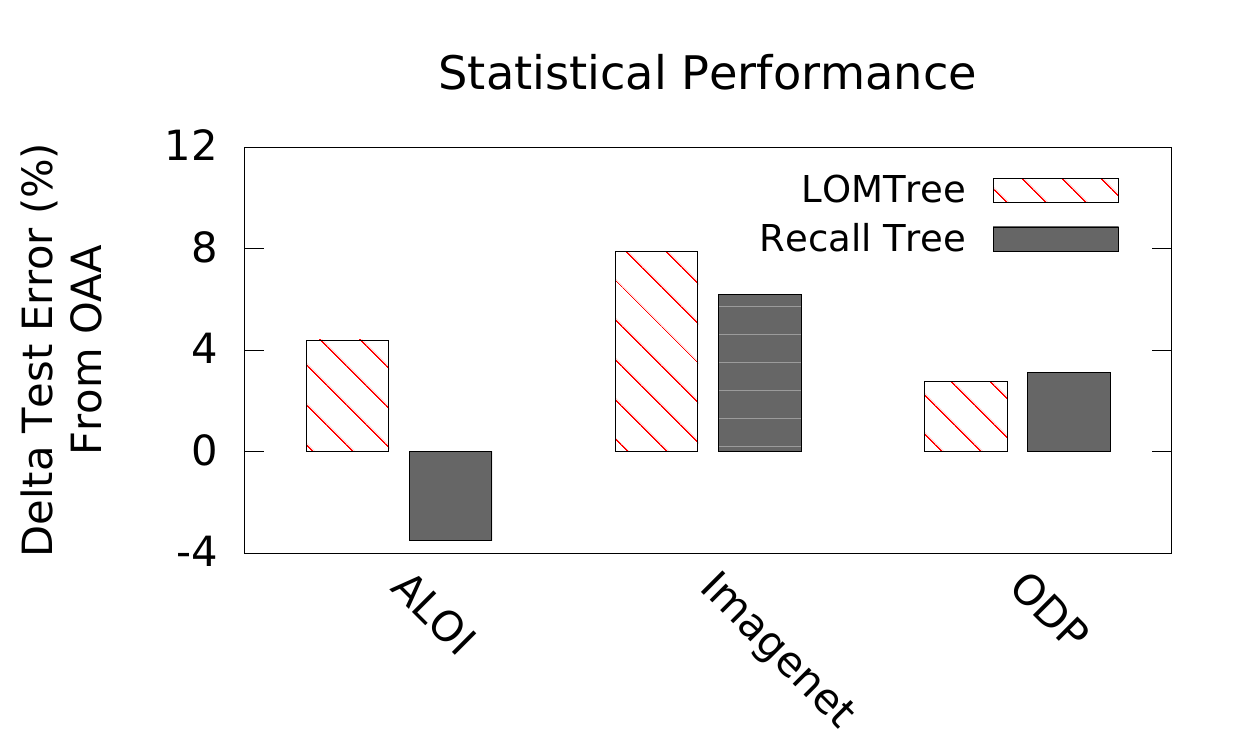}
\end{subfigure}
\begin{subfigure}[t]{0.49\textwidth}
\includegraphics[width=\textwidth]{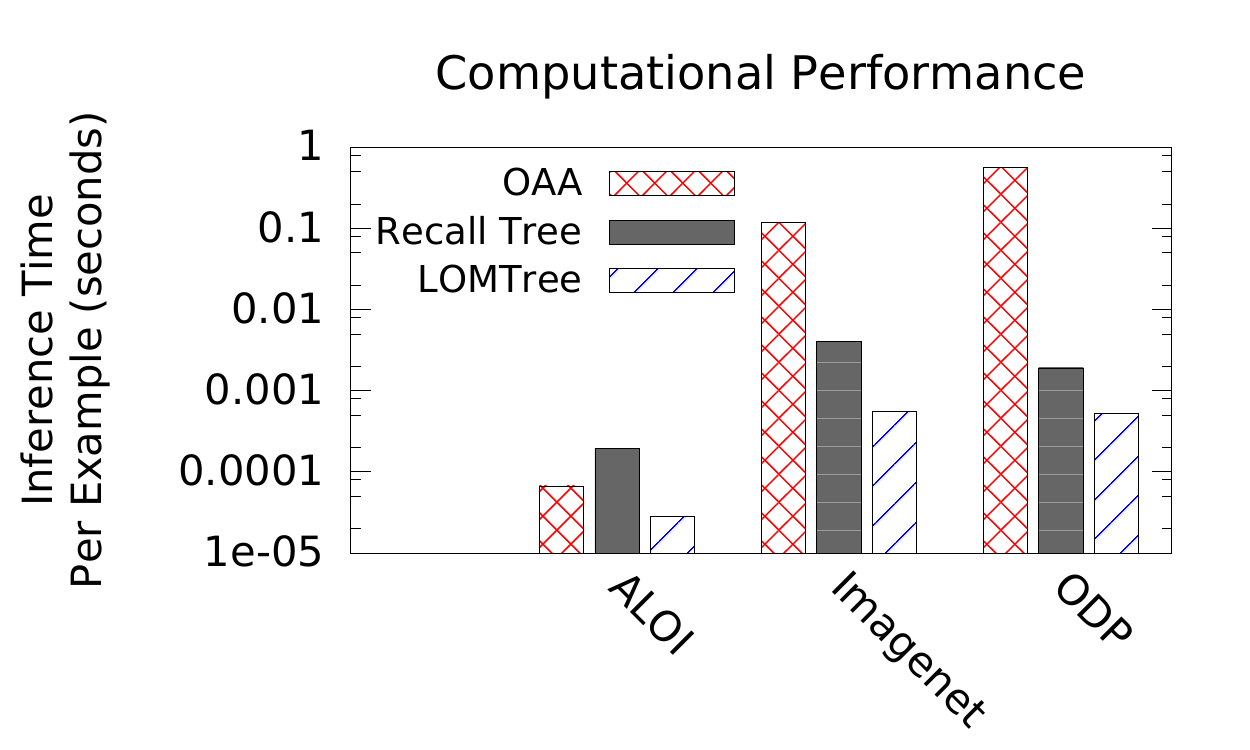}
\end{subfigure}
\caption{Empirical comparison of statistical (left) and computational
(right) performance of Recall Tree against two strong competitors: OAA
(statistically good) and LOMTree (computationally good).  Recall Tree has
$\mathrm{poly} (log)$ dependence upon number of classes (like LOMTree)
but can surpass OAA statistically.}
\label{fig:results}
\end{figure}

In our first set of experiments, we compare Recall Tree with a
strong computational baseline and a strong statistical baseline.
The computational baseline is LOMTree, the only other
online logarithmic-time multiclass algorithm we know of.
The statistical baseline is OAA, whose statistical performance we want to match
(or even exceed), and whose linear computational dependence on the number
of classes we want to avoid.  Details regarding the experimental
methodology are in Appendix~\ref{app:expmethodology}.
Results are summarized in Figure~\ref{fig:results}. 

\paragraph{Comparison with LOMTree}

The Recall Tree uses a factor of 32 less state than the LOMTree which
makes a dramatic difference in feasibility for large scale
applications.  Given this state reduction, the default expectation is
worse prediction performance by the Recall Tree.  Instead, we observe
superior or onpar statistical performance despite the state
constraint.   This typically comes with an additional computational
cost since the Recall Tree evaluates a number of per-class
predictors.

\paragraph{Comparison with OAA}

On one dataset (Aloi) prediction performance is superior to OAA while
on the others it is somewhat worse.  

Computationally OAA has favorable constant factors since it is highly
amenable to vectorization.  Conversely, the conditional execution
pattern of the Recall Tree frustrates vectorization even with example
mini-batching.  Thus on ALOI although Recall Tree does on average 50
hyperplane evaluations per example while OAA does 1000, OAA is
actually faster: larger numbers of classes are required to experience
the asymptotic benefits.  For ODP with $\sim{}10^5$ classes, with
negative gradient subsampling and using 24 cores in parallel, OAA is
about the same wall clock time to train as Recall Tree on a single
core.\footnote{While not yet implemented, Recall Tree can presumably
  also leverage multicore for acceleration.}  Negative gradient
sampling does not improve inference times, which are roughly 300 times
slower for OAA than Recall Tree on ODP.

\subsection{Online Operation}

In this experiment we leverage the online nature of the algorithm to exploit
nonstationarity in the data to improve results.  This is not something that 
is easily done with batch oriented algorithms, or with algorithms that post-process
a trained predictor to accelerate inference.

\begin{figure}
\begin{subfigure}[t]{0.475 \textwidth}
\includegraphics[width=\textwidth]{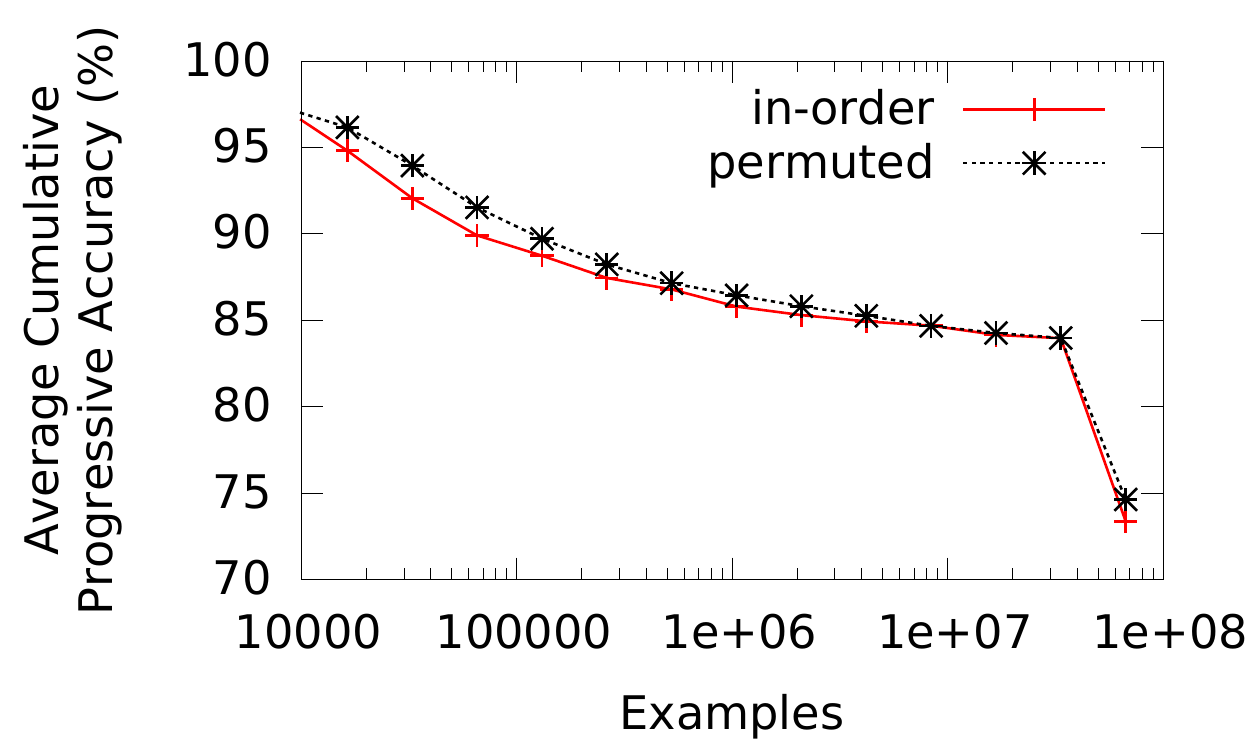}
\caption{When the LTCB dataset is presented in the original order, Recall Tree is able to exploit sequential correlations for improved performance.  After all examples are processed, the average progressive accuracy is 73.3\% vs. 74.6\%.}
\label{fig:online}
\end{subfigure}
\hspace{0.025 \textwidth}
\begin{subfigure}[t]{0.475 \textwidth}
\includegraphics[width=\textwidth]{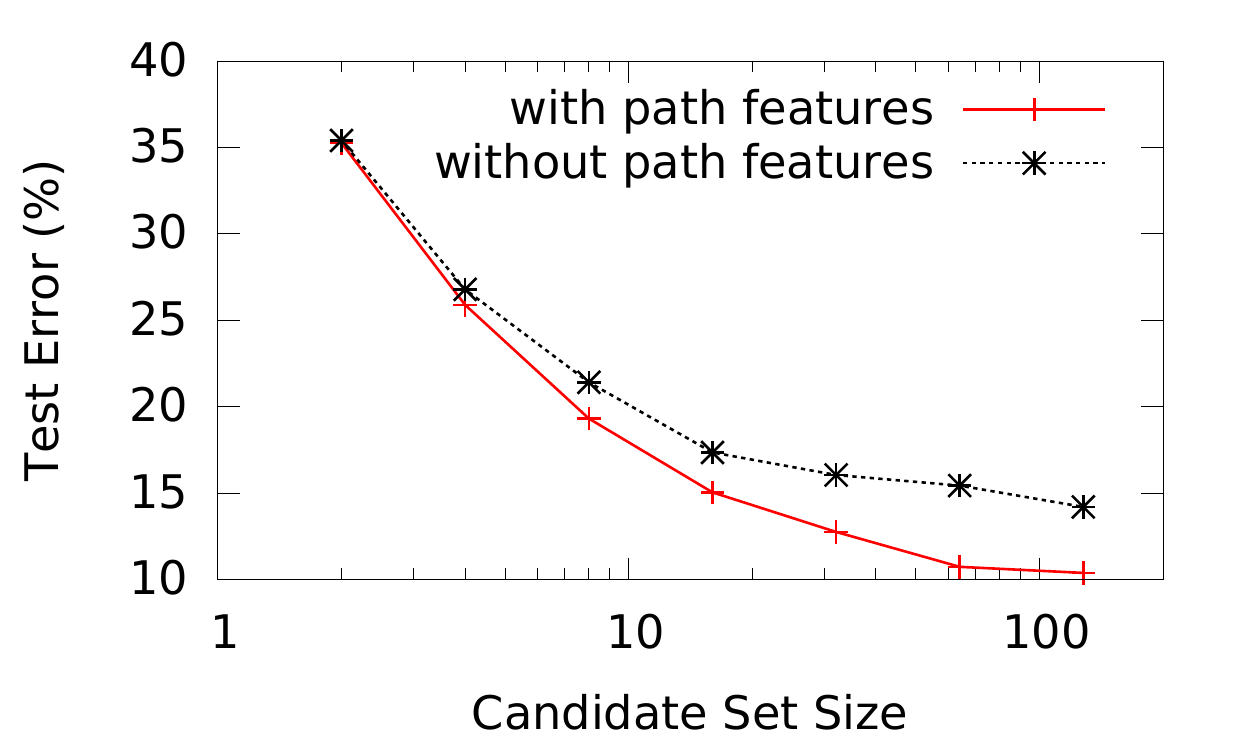}
\caption{Test error on ALOI for various candidate set sizes, with or without path features (all other parameters held fixed).  Using multiple predictors per leaf and including path features improves performance.}
\label{fig:design}
\end{subfigure}
\caption{}
\end{figure}

We consider two versions of LTCB.  In both versions the task is to
predict the next word given the previous 6 tokens.  The difference is
that in one version, the Wikipedia dump is processed in the original
order (``in-order''); whereas in the other version the training data
is permuted prior to input to the learning algorithm (``permuted'').
We assess progressive validation loss~\cite{PV} on the sequence.  The
result in Figure~\ref{fig:online} confirms the Recall Tree is able to
take advantage of the sequentially revealed data; in particular, the
far-right difference in accuracies is significant at a factor
$P<0.0001$ according to an $N-1$ Chi-squared test.

\subsection{Path Features and Multiple Predictors}

Two differences between Recall Tree and LOMTree are the use of multiple
predictors at each tree node and the augmentation of the example with
path features.  In this experiment we explore the impact of these design
choices using the ALOI dataset.

Figure~\ref{fig:design} shows the effect of these two aspects on
statistical performance.  As the candidate set size is increased, test
error decreases, but with diminishing returns.  Disabling path
features degrades performance, and the effect is more pronounced as
the candidate set size increases.  This is expected, as a larger
candidate set size decreases the difficulty of obtaining good recall
(i.e., a good tree) but increases the difficulty of obtaining good
precision (i.e., good class predictors), and path features are only
applicable to the latter. All differences here are significant at a
$P<0.0001$ according to an $N-1$ Chi-squared test, except for when the
candidate set size is $2$, where there is no significant difference.

\subsection{The Empirical Bernstein Bound}

Is the empirical Bernstein bound used helpful?  To test this we
trained on the LTCB dataset with a multiplier on the bound of either
$0$ (i.e. just using empirical recall directly) or $1$.  The results
are stark: with a multiplier of $1$, the test error was $78\%$ while
with a multiplier of $0$ the test error was $91\%$.  Clearly, in the
small samples per class regime this form of direct regularization is
extraordinarily helpful.

\section{Conclusion}

In this work we proposed the Recall Tree, a reduction of multiclass to
binary classification, which operates online and scales
logarithmically with the number of classes. Unlike the
LOMTree~\cite{Lomtree}, we share classifiers among the nodes of the
tree which alleviates data sparsity at deep levels while greatly
reducing the required state.  We also use a tighter analysis which is
more closely followed in the implementation.  These features allow us
to reduce the statistical gap with OAA while still operating many
orders of magnitude faster for large $K$ multiclass datasets.  In the
future we plan to investigate multiway splits in the tree since
$O(\log K)$-way splits does not affect our $O(\textrm{poly}\log K)$
running time and they might reduce contention in the root and nodes
high in the tree.

{\bf Acknowledgements} We would like to thank an anonymous reviewer for
NIPS who spotted an error in the theorem proof and made several good
suggestions for improving it.

\bibliographystyle{plain}
\bibliography{beatoaa}

\newpage
\appendix

\section{Proof of theorem~\ref{thm:boost}}
\label{sec:boostproof}
\begin{proof}
For the fixed tree at timestep $t$ (there have been $t-1$ previous splits)
with a fixed partition function in
the nodes, the weighted entropy of class labels is
$W_{t}=\sum_{\{n\in\mbox{Leaves}\}}f_{n}H_{n}.$

When we split the $t$th node, the weak learning assumption implies entropy decreases by $\gamma$ according to:
$$H_{n}\geq\left(\frac{f_{l}}{f_{n}}H_{l}+\frac{f_{r}}{f_{n}}H_{r}\right)+ \gamma$$
where $\gamma$ is the advantage of the weak learner.  Hence, $$W_{t}-W_{t+1}=f_{n}H_{n}-f_{l}H_{l}-f_{r}H_{r}\geq f_{n} \gamma\,\,\,\,.$$ 

We can bound $f_{n}$ according to $$\max_n f_n \geq \frac{1}{t}$$ which implies
$$W_{t}-W_{t+1}\geq \frac{\gamma}{t}$$.
This can be solved recursively to get:
\begin{align*}
  W_{t+1}  & \leq W_{1} - \gamma \sum_{i=1}^t \frac{1}{i}\\
  & \leq W_{1} - \gamma \left(1+\int_{i=1}^t \frac{1}{i} di \right)\\
  & = W_{1} - \gamma (1+ \ln t)\\
  & = H_{1} - \gamma (1+ \ln t)\\
\end{align*}
where the second inequality follows from bounding each term of the sum
with successive integrals, and $H_1$ is the marginal Shannon entropy
of the class labels.

To finish the proof, we bound the multiclass loss in terms of the
average entropy.  For any leaf node $n$ we can assign the most likely
label, $y = \arg \max_i \pi_{ni}$ so the error rate is $\epsilon_n = 1-\pi_{ny}$. 
\begin{align*}
  W_{t+1}& = \sum_{\{n\in\mbox{Leaves}\}}f_{n}\sum_i \pi_{ni} \ln \frac{1}{\pi_{ni}} \\
    & \geq \sum_{\{n\in\mbox{Leaves}\}}f_{n} \sum_i \pi_{ni} \ln \frac{1}{\pi_{ny}} \\
  & = \sum_{\{n\in\mbox{Leaves}\}}f_{n} \ln \frac{1}{1-\epsilon_n}\\
  & \geq \sum_{\{n\in\mbox{Leaves}\}}f_{n} \epsilon_n = \\
  & = \epsilon
\end{align*}
Putting these inequalities together we have:
\[ \epsilon \leq H_1 - \gamma (1 + \ln t ) \]
\end{proof}

\section{Datasets} \label{app:datasets}

ALOI \cite{geusebroek2005amsterdam} is a color image
collection of one-thousand small objects recorded for scientific
purposes~\cite{geusebroek2005amsterdam}.  We use the same train-test
split and representation as Choromanska et. al.~\cite{Lomtree}.

Imagenet consists of features extracted from intermediate layers of
a convolutional neural network trained on the ILVSRC2012 challenge
dataset. This dataset was originally developed to study
transfer learning in visual tasks~\cite{Oquab14}; more details 
are at \url{http://www.di.ens.fr/willow/research/cnn/}. We utilize a
predictor linear in this representation.

LTCB is the Large Text Compression Benchmark, consisting of the first
billion bytes of a particular Wikipedia dump~\cite{mahoney2009large}.
Originally developed to study text compression, it is now commonly used
as a language modeling benchmark where the task is to predict the next
word in the sequence.  We limit the vocabulary to 80000 words plus a
single out-of-vocabulary indicator; utilize a model linear in the 6
previous unigrams, the previous bigram, and the previous trigram; and
utilize a 90-10 train-test split on entire Wikipedia articles.

ODP\cite{bennett2009refined} is a multiclass dataset derived from the
Open Directory Project. We utilize the same train-test split and labels
from \cite{Lomtree}.  Specifically there is a fixed train-test split of
2:1, the representation of a document is a bag of words, and the class
label is the most specific category associated with each document.

\section{Experimental Methodology} \label{app:expmethodology}

\begin{table}
\caption{Empirical comparison summary.  When OAA training is accelerated using parallelism and gradient subsampling, wall clock times are parenthesized.  Training times are for defaults, i.e., without hyperparameter optimization.  Asterisked LOMTree results are from~\cite{Lomtree}.}
\label{tab:results}
\begin{center}
\begin{tabular}{|c|c|c|c|c|c|}
\hline \multirow{2}{*}{Dataset} & Method & \multicolumn{2}{|c|}{Test Error} & Training Time & Inference Time \\ 
\cline{3-4} & & Default & Tuned &  & per example \\ 
\hline \multirow{3}{*}{ALOI} & OAA & 12.2\% & 12.1\% & 571s & 67$\mu$s\\
& Recall Tree & 11.4\% & 8.6\% & 1972s & 194$\mu$s \\
& LOMTree & 21.4\% & $16.5\%^*$ & 112s & 28$\mu$s \\ 
\hline \multirow{3}{*}{Imagenet} & OAA & 84.7\% & 82.2\% & 446d (20.4h) & 118ms \\
& Recall Tree & 91.1\% & 88.4\% & 71.4h & 4ms \\
& LOMTree & 96.7\% & $90.1\%^*$ & 14.0h & 0.56ms \\ 
\hline \multirow{3}{*}{LTCB} & OAA & 78.7\% & 76.8\% & 764d (19.1h) & 3600$\mu$s \\
& Recall Tree & 78.0\% & 77.6\% & 4.8h & 76$\mu$s \\
& LOMTree & 78.4\% & - & 4.3h & 51$\mu$s \\ 
\hline \multirow{3}{*}{ODP} & OAA & 91.2\% & 90.6\% & 133d (1.3h) & 560ms \\
& Recall Tree & 96.2\% & 93.8\% & 1.5h & 1.9ms \\
& LOMTree & 95.4\% & $93.5\%^*$ & 0.6h & 0.52ms \\ 
\hline
\end{tabular}
\end{center}
\end{table}

\paragraph{Default Performance Methodology} 
\begin{wraptable}{r}{0.535\textwidth}
\vspace{-8pt}
\begin{tabular}{|c|c|c|} \hline
Algorithm & Parameter & Default Value \\ \hline
\multirow{2}{*}{Binary} & Learning Rate & 1 \\ \cline{2-3}
                        & Loss & logistic \\ \hline
\multirow{2}{*}{Recall Tree} & Max Depth & $\log_2 (\mbox{\#classes})$ \\ \cline{2-3}
                             & Num Candidates & $4 \log_2 (\mbox{\#classes})$ \\ \cline{2-3}
                             & Depth Penalty ($\lambda$) & 1 \\ \hline
%\multirow{2}{*}{LOMTree} & Swap Resistance & 1 \\ \cline{2-3}
%                         & Tree Nodes & \#classes \\ \hline
\end{tabular}
\caption{Algorithm hyperparameters for various algorithms.  ``Binary'' refers to hyperparameters inherited via reduction to binary classification.}
\label{tab:hyper}
\vspace{-24pt}
\end{wraptable}
Hyperparameter selection can be computationally burdensome for large
data sets, which is relevant to any claims of decreased training
times. Therefore we report results using the default values indicated
in Table~\ref{tab:hyper}.  For the larger data sets (Imagenet, ODP), we
do a single pass over the training data; for the smaller data set (ALOI),
we do multiple passes over the training data, monitoring a 10\% held-out
portion of the training data to determine when to stop optimizing.

\paragraph{Tuned Performance Methodology} For tuned performance, we use
random search over hyperparameters, taking the best result over 59 probes.
For the smaller data set (ALOI), we optimize validation error on
a 10\% held-out subset of the training data.  For the larger data sets
(Imagenet, ODP), we optimize progressive validation loss on the
initial 10\% of the training data.  After determining hyperparameters we
retrain with the entire training set and report the resulting test error.

When available we report published LOMtree results, although they utilized
a different method for optimizing hyperparameters.

\paragraph{Timing Measurements}  All timings are taken from the same 24
core xeon server machine.  Furthermore, all algorithms are implemented in
the Vowpal Wabbit toolkit and therefore share file formats, parser, and
binary classification base learner implying differences are attributable
to the different reductions.  Our baseline OAA implementation is mature
and highly tuned: it always exploits vectorization, and furthermore can
optionally utilize multicore training and negative gradient subsampling
to accelerate training.  For the larger datasets these latter features
were necessary to complete the experiments: estimated unaccelerated
training times are given, along with wall clock times in parenthesis.

\end{document}